\documentclass{article}
\usepackage{spconf,amsmath,graphicx,hyperref}
\usepackage{amssymb}
\usepackage{booktabs} 
\usepackage{multirow} 

\title{Code-Switching Detection via Language-Selective Dimensions \\ in Speech Foundation Models}
\name{Jie Chi, Zakaria Aldeneh, 
Masha Fedzechkina, Maureen de Seyssel, Natalie Schluter}
\address{Apple}
\begin{document}
\ninept
\maketitle
\begin{abstract}
Code-switching (CS), the alternation between languages within an utterance, is common in multilingual speech but underrepresented in pretraining corpora. Increasing CS coverage requires finding relevant utterances in large speech collections so that they can be included in training data. One way to find such utterances is to transcribe the speech and detect language alternation in the text, but this depends on obtaining accurate multilingual transcriptions. We propose a lightweight method that detects CS directly from speech. It uses frame-level language information in pretrained speech encoders to determine whether both languages are present within an utterance. Specifically, we identify language-selective hidden-state dimensions using monolingual recordings and aggregate their frame-level activations into a CS prediction. With a small CS set used to tune a decision rule, we show that the Whisper-small encoder achieves a mean F1 of $0.798$ across 14 language pairs in CS-FLEURS, compared with $0.776$ for a CNN and $0.751$ for a HuBERT linear baseline trained for monolingual language identification. Our findings suggest that pretrained speech encoders already encode the information needed to identify CS utterances, and they provide a lightweight tool for building more representative multilingual corpora.

\end{abstract}
\section{Introduction}
\label{sec:intro}
Multilingual speech corpora cover many languages but often contain few code-switched (CS) utterances. As a result, models trained on these corpora have limited exposure to mixed-language speech. Recent work shows that training or adapting ASR systems with CS data improves recognition of such speech \cite{liu2024aligning,nguyen2025asyncswitch,yeo2026improving}. A recent study introducing CS-YODAS, a dataset of naturally occurring CS speech mined from the large multilingual YODAS speech corpus, also reports gains in spoken language identification when this speech is added to training \cite{yan2026csyodas}.

Existing corpora such as SEAME \cite{lyu2010seame}, the Miami corpora \cite{deuchar2014building}, and the South African soap-opera corpus \cite{van_der_westhuizen2018corpus} provide natural CS speech, but the amount available for training is limited by their size, language-pair coverage, and the proportion of recordings that contain a switch. For example, only 8\% of utterances in the Miami corpus are code-switched. Even in SEAME, where some interview questions were designed to elicit code-switching, the reported proportion is 52\% \cite{chi-bell-2024-analyzing}. Therefore, larger multilingual CS benchmarks have relied mainly on read or synthetic speech \cite{diwan2021mucs,hussein2024speech,yan2025csfleurs}. Mining CS utterances from larger speech collections could provide more natural examples across languages and domains.

Mining CS utterances from a large speech collection requires identifying which recordings contain CS. CS-YODAS \cite{yan2026csyodas} shows that natural CS speech can be mined from web audio using a pipeline that uses ASR transcripts, a 14B-parameter language model, and bilingual human feedback. Having access to a model that can detect CS directly from speech can be beneficial as a pre-filter before running more expensive parts of the pipeline. We therefore investigate whether CS can be detected directly from speech with a lightweight binary detector.

Prior work has shown that frame-level hidden representations of pretrained speech models encode information about spoken language \cite{de_seyssel2022probing}. In this work, we ask whether this frame-level information can be used to detect whether code-switching has occurred within a given utterance. Our approach proceeds in two steps: we first identify hidden-state dimensions in a pretrained speech encoder whose activations differ between the target languages, using a small set of monolingual recordings; then, at test time, we monitor these same dimensions to predict whether a switch in language occurs within a given utterance. We study the utility of this approach and show that it yields an effective CS detector.

\section{Related work}
\label{sec:related work}

CS can be detected from speech without first transcribing it. Systems trained on labelled monolingual and CS speech have been evaluated for utterance-level detection on three English--Indian-language pairs, as well as for frame-level language identification \cite{rangan2020spectral}. CS language identification has also been studied on English--Mandarin child-directed speech using CNN--GRU models and ASR-assisted pretraining \cite{li2023simple}. These systems are developed and evaluated for specific language pairs and speech settings. Applying a trained CS detector to a new pair may require labelled CS examples, which can be difficult to obtain before mining begins.

Language diarization offers another way to mine samples with CS by assigning language labels to temporal regions of speech. An end-to-end approach adapted from speaker diarization trains on CS utterances with language labels at 200 ms intervals, but reports a substantial performance drop from synthetic to recorded speech, partly attributed to differences in segment duration \cite{mishra2024implicit}. Another approach fine-tunes speech representations on 14.3 hours of annotated South African soap-opera speech spanning English and four other languages; approximately 4.5 hours of CS speech are in its training set \cite{frost2022finetuned}. The DISPLACE challenges evaluate diarization on multilingual conversational speech \cite{baghel2023displace,kalluri2024second}, with one strong system fine-tuning its language representations on in-domain data \cite{kalda2024taltech}. These results show that language boundaries can be recovered from speech, while also illustrating the role of annotated CS data and recording conditions. We examine whether likely CS utterances can instead be identified using dimensions selected from monolingual speech and a small labelled CS set for tuning. We test this approach across language pairs and on conversational speech from a second domain.

\section{Method}
\label{sec:method}
In this section, we describe how we identify language-selective dimensions in a pretrained speech model and use this information to detect CS. We first measure how each dimension's activations differ between two languages in monolingual speech. We then test whether the same dimensions are selective across language pairs. Finally, we use their frame-level activations to predict whether an utterance contains CS. We use monolingual speech to select dimensions and estimate language centroids, and code-switched speech to evaluate the detector and tune its decision rule. The corpora, language pairs, and data splits used for each step are described in Section \ref{sec:experimental_setup}
\subsection{Language-Selective Dimensions}
\label{sec:dimensions}
If language information in a multi-lingual speech model is concentrated in a small number of dimensions, those dimensions may provide a compact signal for detecting both languages within an utterance. We therefore measure how strongly each dimension separates monolingual speech from two languages. Let $\mathbf{h}_t^{(l)} \in \mathbb{R}^d$ denote the hidden state at frame $t$ after Transformer block $l$. We analyse its individual dimensions, corresponding to coordinates of the encoder residual stream.\footnote{We also evaluate FFN intermediate activations, taken after the nonlinearity and before the second feed-forward projection; e.g., on Whisper-small, FFN intermediate activations are 3072-dimensional while residual-stream activations are 768-dimensional. The residual stream gives better code-switching detection, so we report residual-stream results throughout.} For a language pair $(A,B)$, we rank dimensions using the absolute standardised difference between their mean activations, Cohen's~$d$:
\begin{equation}
d_i =
\frac{|\mu_i^A-\mu_i^B|}
{\sqrt{\left((\sigma_i^A)^2+(\sigma_i^B)^2\right)/2}},
\label{eq:cohend}
\end{equation}
where $\mu_i^A$ and $\sigma_i^A$ are the mean and standard deviation of dimension $i$ across frames from language $A$, and likewise for $B$. A larger $d_i$ indicates a larger difference in mean activation relative to within-language variation. We use the maximum $d_i$ in each layer to choose one layer per encoder, then rank dimensions within that layer for each language pair. We do not compare $d_i$ values across models.
\subsection{Cross-Pair Recurrence}
\label{sec:recurrence_def}

A dimension that is selective for one language pair may not be selective for others. We therefore measure how often the same dimensions rank highly across pairs. At the selected layer, we rank dimensions by Cohen's~$d$ for each pair and record the top $K$.

For dimension $i$, we define its \textit{cross-pair recurrence} as
\begin{equation}
U_i =
\frac{1}{|\mathcal{P}|}
\sum_{(A,B)\in\mathcal{P}}
\mathbb{I}\left[\operatorname{rank}_{A,B}(i) \leq K\right],
\label{eq:recurrence}
\end{equation}
where $\mathcal{P}$ is the set of language pairs and $K$ is the number of dimensions in the top $1\%$, rounded to the nearest integer. We report $\max_i U_i$, the recurrence of the most frequently selected dimension. Since we take the maximum over all dimensions, the chance level is higher than $K/d$. We estimate it by Monte Carlo simulation, randomly selecting $K$ dimensions for each pair and computing the maximum recurrence each time.

\subsection{Code-Switching Detection}
\label{sec:cs_method}
We use the selected dimensions to obtain a language score for each frame. Monolingual speech establishes the typical activations for languages $A$ and $B$ along a language axis; frames closer to each language's centroid on this axis receive scores in opposite directions. We then test whether the utterance contains sufficient local evidence for each language.

For a language pair $(A,B)$, we estimate language centroids $\boldsymbol{\mu}_A$ and $\boldsymbol{\mu}_B$ from monolingual speech. Given a set of selected dimensions $\mathcal{K}$, we define a unit-length axis between the centroids:
\begin{equation}
\mathbf{v} =
\frac{(\boldsymbol{\mu}_A-\boldsymbol{\mu}_B)[\mathcal{K}]}
{\|(\boldsymbol{\mu}_A-\boldsymbol{\mu}_B)[\mathcal{K}]\|_2},
\end{equation}
and project their midpoint onto this axis:
\begin{equation}
c_{\mathrm{mid}} =
\frac{1}{2}
(\boldsymbol{\mu}_A+\boldsymbol{\mu}_B)[\mathcal{K}]
\cdot \mathbf{v}.
\end{equation}
The score for frame $t$ is
\begin{equation}
s_t =
\mathbf{h}_t[\mathcal{K}] \cdot \mathbf{v}
-c_{\mathrm{mid}},
\label{eq:score}
\end{equation}
where positive scores point towards the $A$ centroid and negative scores towards the $B$ centroid.

We use a 0.5-second sliding window with a 0.1-second stride. Within each window, $p_A$ and $p_B$ are the fractions of frames with positive and negative scores, respectively. We classify an utterance as code-switched if at least $n$ windows satisfy $p_A \geq \theta$ and at least $n$ windows satisfy $p_B \geq \theta$. These need not be consecutive windows. We tune $\theta$ and $n$ on a small labelled set, as described in Section \ref{sec:experimental_setup}.

\section{Experimental Setup}
\label{sec:experimental_setup}

\subsection{Data}

We use three speech corpora. \textbf{SEAME} \cite{lyu2010seame} contains approximately 192 hours of spontaneous Mandarin--English speech from 156 speakers in Singapore and Malaysia. We use monolingual training utterances to select encoder layers, estimate English and Mandarin centroids, and train the three baselines. We evaluate CS detection on its two official development sets, \textit{dev\_man} (Mandarin-dominant) and \textit{dev\_sge} (Singapore English-dominant).

\textbf{FLEURS} \cite{conneau2023fleurs} contains read speech in 102 languages. We use monolingual training utterances from English and the 14 non-English languages in our CS-FLEURS evaluation. These 15 languages form 105 pairs for the cross-pair recurrence analysis. The same training utterances are used for pair-specific dimension selection, centroid estimation, and baseline training.

\textbf{CS-FLEURS} \cite{yan2025csfleurs} contains read and synthetic CS speech, covering 113 language pairs and 52 languages in total. We evaluate only on its read-speech portion, using 14 English--X pairs.

For both evaluation corpora, CS utterances are used to tune the detection rule and compute CS detection F1. Neither corpus contributes CS utterances to dimension selection or centroid estimation.

\subsection{Pretrained Models}

We study three pretrained multilingual speech encoders with different training objectives and language coverage. \textbf{Whisper-small} \cite{radford2023whisper} is trained with supervised multitask learning that includes predicting a language token. It covers approximately 100 languages and has a hidden size of $d=768$. \textbf{mHuBERT-147} \cite{nguyen2023mhubert} uses self-supervised masked prediction on speech from 147 languages, without an explicit language identification objective ($d=768$). \textbf{MMS-1B} \cite{pratap2023mms} uses wav2vec2.0-style contrastive pretraining on 1,406 languages ($d=1{,}280$).

\subsection{Layer and Dimension Selection}
For each model, we select one encoder layer using monolingual SEAME training speech, so the layer is not chosen to maximise cross-pair recurrence on FLEURS.  We sample 300 English and 300 Mandarin utterances, compute Cohen's~$d$ for every dimension at every layer, and select the layer with the highest peak selectivity. Three independent samples of 300 utterances per language selected the same layer.

We keep this layer fixed for the remaining analyses. To measure cross-pair recurrence, we sample 300 training utterances from each of the 15 FLEURS languages and compute Cohen's~$d$ for all 105 pairs. We then rank dimensions by their recurrence score (Eq.~\ref{eq:recurrence}).

For CS detection, we compare two ways of selecting $K$ dimensions. \textbf{Recurring dimensions} uses the dimensions with the highest cross-pair recurrence, with the same indices for every target pair. \textbf{Pair-specific dimensions} uses the dimensions with the highest Cohen's~$d$ for each target pair, selected from its monolingual training speech. We set $K{=}8$ for the two models with $d{=}768$ and $K{=}13$ for MMS-1B with $d{=}1{,}280$. For both strategies, language centroids are estimated from monolingual speech in the corpus used for the corresponding evaluation.

\subsection{Baselines}

We compare with three classifiers trained to identify the target languages from monolingual speech. Each produces frame-level language predictions at inference, which we aggregate into a CS decision using the same sliding-window rule as our method.

\textbf{CNN} is a four-layer 1D convolutional network operating on 80-dimensional log-Mel features extracted with a 25\,ms window and 10\,ms shift. Each layer has 256 channels and a kernel size of 5, followed by batch normalization and ReLU. The model is trained with utterance-level language labels using cross-entropy and Adam ($\mathrm{lr}{=}10^{-3}$) for 30 epochs. \textbf{HuBERT linear} is a two-class linear head trained on mean-pooled representations from frozen HuBERT-base ($d{=}768$), using the same optimizer and number of epochs. At inference, the head is applied to each frame. \textbf{Full-layer probe} uses the same training and inference procedure with frozen Whisper-small layer~8 representations ($d{=}768$). It tests whether a classifier trained on the full layer performs better than using a small set of selected dimensions.\footnote{We also evaluated a linear head on mHuBERT-147 representations, but it performed worse than the HuBERT-base.  Both  HuBERT linear and full-layer probe achieve near-perfect monolingual LID accuracy on FLEURS }

For CS-FLEURS, we train each baseline separately for each language pair on monolingual FLEURS training utterances. For SEAME, we train them on monolingual English and Mandarin utterances from its training set. We balance the two languages by subsampling each to $\min(n_A,n_B)$ utterances. On SEAME, this leaves over 22,000 training utterances per language for each baseline, compared with 300 per language for dimension selection.

\subsection{Code-Switching Evaluation}
We report utterance-level F1 for CS detection. We select the threshold $\theta$ and minimum window count $n$ by grid search over $\theta \in \{0.05, 0.10, \ldots, 0.95\}$ and $n \in \{1,\ldots,5\}$. We report two settings. \textbf{Oracle} selects the parameters that maximise F1 on the full evaluation set. \textbf{Calibration} tests parameter selection with limited labelled data. For CS-FLEURS, we select $(\theta,n)$ on a random 10\% of evaluation utterances and evaluate on the remaining 90\%. For SEAME, we select $(\theta,n)$ on 200 randomly sampled CS training utterances and an equal number of monolingual training utterances, then evaluate on the development sets. We repeat each calibration procedure 200 times and report mean F1 and standard deviation.

\begin{table*}[t]
  \centering
  \caption{Utterance-level CS detection F1. Oracle parameters are selected on the evaluation set; calibrated parameters are selected from limited labelled data. Calibrated results show mean and standard deviation over 200 repetitions.}
  \label{tab:cs_results}
  \small
  \setlength{\tabcolsep}{4pt}
  \begin{tabular}{llc rrr rrr}
    \toprule
    & & & \multicolumn{3}{c}{CS-FLEURS} & \multicolumn{3}{c}{SEAME\_combined} \\
    \cmidrule(lr){4-6} \cmidrule(lr){7-9}
    & & & & \multicolumn{2}{c}{Calibrated} & & \multicolumn{2}{c}{Calibrated} \\
    \cmidrule(lr){5-6} \cmidrule(lr){8-9}
    Method & Model & $K$ & Oracle & Mean & Std & Oracle & Mean & Std \\
    \midrule
    Pair-specific, $K=1\%$      & Whisper-small & 8  & \textbf{0.798} & 0.784 & .021 & 0.734 & 0.733 & .008 \\
    Pair-specific, $K=1\%$      & mHuBERT-147   & 8  & 0.756 & 0.737 & .022 & 0.732 & 0.731 & .009 \\
    Pair-specific, $K=1\%$      & MMS-1B        & 13 & 0.762 & 0.745 & .020 & \textbf{0.758} & \textbf{0.755} & .006 \\
    Recurring dims.\ (all pairs) & Whisper-small & 8  & 0.790 & 0.774 & .020 & 0.736 & 0.734 & .011 \\
    Recurring dims.\ (all pairs) & mHuBERT-147   & 8  & 0.697 & 0.681 & .021 & 0.737 & 0.735 & .009 \\
    Recurring dims.\ (all pairs) & MMS-1B        & 13 & 0.735 & 0.720 & .019 & 0.733 & 0.732 & .009 \\
    Recurring dims., leave-X-out    & Whisper-small & 8  & 0.797 & 0.783 & .020 & \multicolumn{3}{c}{N/A} \\
    Recurring dims., leave-both-out & Whisper-small & 8  & 0.795 & 0.781 & .020 & \multicolumn{3}{c}{N/A} \\
    \midrule
    CNN classifier           & ---           & --- & 0.776 & 0.760 & .017 & 0.753 & 0.740 & .006\\
    HuBERT linear classifier & HuBERT        & --- & 0.751 & 0.733 & .019 & 0.772 & 0.769 & .009 \\
    Full-layer probe         & Whisper-small & --- & 0.786 & 0.768 & .019 & 0.721 & 0.720 & .008 \\
    \bottomrule
  \end{tabular}
\end{table*}

\section{Results}
\label{sec:results}

We first examine whether language-selective dimensions recur across pairs and recording domains. If they do, the same dimensions may provide language information for CS detection beyond the pair on which they were identified. We then test this possibility through utterance-level CS detection, per-pair results, and transfer to held-out pairs.

\subsection{Language-Selective Dimensions Recur Across Pairs}
\label{sec:recurrence}

We first ask whether dimensions selected for one language pair also rank highly for others. If so, the same dimensions may support CS detection across pairs, rather than requiring an unrelated set for each one. Across the 105 FLEURS pairs, the most frequently selected dimension ranks in the top $1\%$ for 55.2\% of pairs in Whisper-small, 26.7\% in mHuBERT-147, and 16.2\% in MMS-1B. The corresponding Monte Carlo chance levels are 5.3\% ($\pm 0.7\%$) for the 768-dimensional models and 5.5\% ($\pm 0.7\%$) for MMS-1B. Recurrence is highest in Whisper-small, although differences in objective, language coverage, and training data prevent us from attributing this result to one aspect of pretraining.

\subsection{Some Dimensions Remain Selective Across Domains}
\label{sec:universal}

We examine the recurring Whisper-small dimensions more closely. At layer8, dimensions 75, 332, and 704 rank in the top eight for 55.2\%, 51.4\%, and 51.4\% of the 105 pairs, respectively; dimensions 398 and 466 also recur frequently. This raises a further question for CS detection: \textit{do dimensions selected from read speech remain selective in conversational speech?}

For English--Mandarin, per-dimension Cohen's $d$ rankings on FLEURS and SEAME have a weak Spearman correlation of $\rho{=}0.251$. Three of the eight most recurring dimensions nevertheless rank in the top eight in both corpora. Dimension 332 ranks first in SEAME, while dimensions 75 and 398 rank in the top seven in both. Other dimensions transfer less consistently: dimension 704 ranks 24th in FLEURS and 26th in SEAME for this pair, whereas dimension 466 falls from 118th to 741st. Thus, some highly recurring dimensions remain selective across domains, but recurrence alone does not guarantee domain stability.

\subsection{Selected Dimensions Detect CS Across Domains}
\label{sec:cs}

Table~\ref{tab:cs_results} reports utterance-level CS detection F1. On CS-FLEURS, Whisper-small reaches the highest mean oracle F1, $0.798$, using eight pair-specific dimensions. It exceeds the CNN ($0.776$), HuBERT linear ($0.751$), and full-layer Whisper-small probe ($0.786$). mHuBERT-147 and MMS-1B reach $0.756$ and $0.762$, respectively.

With $\theta$ and $n$ selected from 10\% of labelled CS-FLEURS utterances, Whisper-small retains mean F1 of $0.784\pm0.021$, compared with $0.798$ under oracle selection. This calibration subset is small for most pairs. For example, Italian--English has only 176 CS utterances in total, so a random 10\% sample contains about 18 CS examples on average.

For SEAME, Table~\ref{tab:cs_results} reports F1 on the combined \textit{dev\_man} and \textit{dev\_sge} sets. MMS-1B has the highest oracle F1 among the dimension methods ($0.758$), while HuBERT linear scores highest overall ($0.772$). The calibrated scores are close to the corresponding oracle scores. In the split-wise results, F1 is always lower on \textit{dev\_sge} than on \textit{dev\_man}

\subsection{CS Detection Varies Across Language Pairs}
Table~\ref{tab:per_pair} reports oracle F1 for each CS-FLEURS pair. Whisper-small ranges from $0.754$ on German--English to $0.870$ on Telugu--English and has the highest score on 11 of the 14 pairs. mHuBERT-147 leads on Portuguese--English ($0.862$) and Spanish--English ($0.890$), while MMS-1B leads on Telugu--English with the highest single-pair score ($0.918$). Whisper-small has the highest mean F1, but the strongest model differs by pair.
\begin{table}[t]
  \centering
  \caption{Utterance-level CS detection F1 for each CS-FLEURS language pair using pair-specific dimensions and oracle parameter selection.}
  \label{tab:per_pair}
  \small
  \setlength{\tabcolsep}{4pt}
  \begin{tabular}{lrrr}
    \toprule
    Pair & Whisper-small & mHuBERT-147 & MMS-1B \\
    \midrule
    ara-eng & 0.832 & 0.806 & 0.824 \\
    ces-eng & 0.793 & 0.709 & 0.674 \\
    cmn-eng & 0.762 & 0.693 & 0.728 \\
    deu-eng & 0.754 & 0.672 & 0.677 \\
    fra-eng & 0.792 & 0.712 & 0.727 \\
    hin-eng & 0.846 & 0.762 & 0.835 \\
    ita-eng & 0.779 & 0.722 & 0.735 \\
    jpn-eng & 0.785 & 0.743 & 0.773 \\
    kor-eng & 0.800 & 0.693 & 0.785 \\
    por-eng & 0.774 & \textbf{0.862} & 0.800 \\
    rus-eng & 0.827 & 0.770 & 0.785 \\
    slk-eng & 0.781 & 0.701 & 0.704 \\
    spa-eng & 0.775 & \textbf{0.890} & 0.703 \\
    tel-eng & 0.870 & 0.855 & \textbf{0.918} \\
    \midrule
    \textbf{Mean} & \textbf{0.798} & 0.756 & 0.762 \\
    \bottomrule
  \end{tabular}
\end{table}

\subsection{Dimension Selection Transfers to Held-Out Pairs}
\label{sec:loo}

We test whether dimension indices must be selected using the target pair. For each of the 14 CS-FLEURS pairs, we rank dimensions by recurrence over a subset of the 105 FLEURS pairs using Whisper-small. \textbf{Leave-X-out} excludes the non-English target language, leaving 91 pairs for dimension selection. \textbf{Leave-both-out} excludes both target languages, leaving 78 pairs. In both conditions, we estimate centroids from monolingual speech in the target languages; only the dimension indices are selected without them.

Leave-X-out and leave-both-out achieve mean oracle F1 of $0.797$ and $0.795$, compared with $0.798$ for pair-specific selection. Their calibrated scores are also similar ($0.783$ and $0.781$, compared with $0.784$). On these 14 pairs, selecting dimensions without the target pair has little effect on CS detection F1, provided that monolingual speech from the target languages is available for centroid estimation.

\section{Conclusion}

We present a lightweight CS detector intended for selecting candidate utterances from speech corpora. It identifies language-selective dimensions using only monolingual speech. With approximately $1\%$ of one Whisper-small encoder layer, it achieves oracle F1 of $0.798$ on CS-FLEURS, exceeding the CNN and HuBERT linear baselines. Its F1 on conversational SEAME speech is $0.734$. The most frequently selected Whisper-small dimension ranks in the top $1\%$ for $55.2\%$ of language pairs, compared with $5.3\%$ under random selection. Excluding both target languages when selecting dimension indices lowers CS-FLEURS F1 by $0.003$, while target-language monolingual speech is still used to estimate centroids. However, these results are based on benchmark datasets; we have not yet tested the detector as part of a mining pipeline on unlabelled, in-the-wild audio. It also requires a small labelled CS set to tune its decision rule.  Future work will test candidate selection in that setting and investigate calibration with no CS labels.


\bibliographystyle{IEEEbib}
\bibliography{refs}

\end{document}